\documentclass[runningheads]{llncs}

\usepackage[year=2024,ID=2686]{eccv}

\usepackage{eccvabbrv}

\usepackage{bbding}
\usepackage{graphicx}
\usepackage{booktabs}

\usepackage[accsupp]{axessibility}  

\usepackage[pagebackref,breaklinks,colorlinks,citecolor=eccvblue]{hyperref}

\usepackage{orcidlink}

\begin{document}

\title{DISCO: Embodied Navigation and Interaction via Differentiable Scene Semantics and Dual-level Control} 

\titlerunning{DISCO}

\author{
Xinyu Xu\inst{1}\orcidlink{0009-0002-1847-9913} 
Shengcheng Luo\inst{1} \orcidlink{0000-0002-2671-1236}
Yanchao Yang\inst{2}\orcidlink{0000-0002-2447-7917} 
Yong-Lu Li\inst{1}$^{\dag}$\orcidlink{0000-0003-0478-0692}
Cewu Lu\inst{1}$^{\dag}$\orcidlink{0000-0003-1533-8576}
}

\authorrunning{X. Xu et al.}

\institute{Shanghai Jiao Tong University \\
\email{\{xuxinyu2000,yonglu\_li,lucewu\}@sjtu.edu.cn woodroof1998@gmail.com}
\and The University of Hong Kong \\
\email{yanchaoy@hku.hk}
}

\maketitle
\let\thefootnote\relax\footnotetext{$^\dag$ Corresponding authors.}
\begin{abstract}

Building a general-purpose intelligent home-assistant agent skilled in diverse tasks by human commands is a long-term blueprint of embodied AI research, which poses requirements on task planning, environment modeling, and object interaction.
In this work, we study primitive mobile manipulations for embodied agents, \ie how to navigate and interact based on an instructed verb-noun pair.
We propose \textbf{DISCO}, which features non-trivial advancements in contextualized scene modeling and efficient controls.
In particular, DISCO incorporates differentiable scene representations of rich semantics in object and affordance, which is dynamically learned on the fly and facilitates navigation planning. 
Besides, we propose dual-level coarse-to-fine action controls leveraging both global and local cues to accomplish mobile manipulation tasks efficiently. 
DISCO easily integrates into embodied tasks such as embodied instruction following.
To validate our approach, we take the ALFRED benchmark of large-scale long-horizon vision-language navigation and interaction tasks as a test bed.
In extensive experiments, we make comprehensive evaluations and demonstrate that DISCO outperforms the art by a sizable +8.6\% success rate margin in unseen scenes, even without step-by-step instructions.
Our code is publicly released at \href{https://github.com/AllenXuuu/DISCO}{https://github.com/AllenXuuu/DISCO}.

\end{abstract}

\section{Introduction}
\label{sec:intro}

\begin{figure}[t]
\centering
\includegraphics[width=0.7\textwidth]{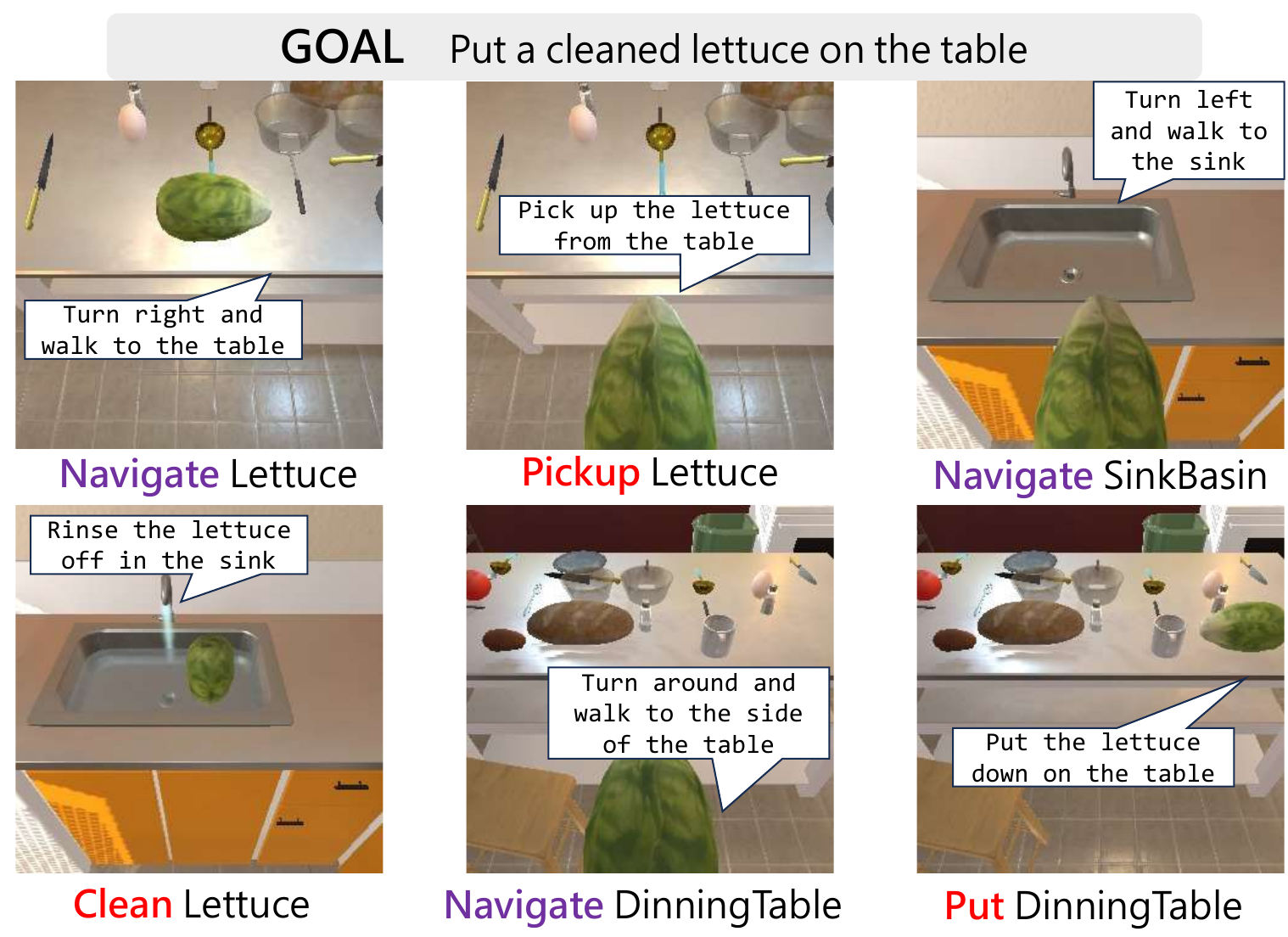}
\caption{An example of vision-language navigation and interaction task in ALFRED~\cite{alfred}. An agent is given a goal directive and step-by-step instructions to perform mobile manipulation of multiple subgoals. Our work can omit step-by-step instructions.}
\label{fig:teaser}
\end{figure}
 
Recent years have witnessed a huge effort~\cite{alfred,rxr,r2r,behavior} in developing embodied agents to perform everyday household tasks in indoor environments. 
However, accomplishing long-horizon household tasks in the unstructured real world by human commands remains challenging for modern robots.
Numerous fundamental capabilities are essential for such a comprehensive robotic application, encompassing multimodal understanding, decomposing long-term objectives, perceiving the environment, and taking actions.
\cref{fig:teaser} gives an intuitive case. 
The agent parses human directives to make corresponding plans. 
Then, it perceives the surroundings to localize semantic entities, navigates to desired waypoints, and interacts with objects.

We center on \textbf{primitive tasks} of mobile manipulations in this work, which necessitates fundamental capability to navigate and interact based on an instructed \textit{verb-noun pair}, \eg \textit{Pickup Lettuce}.
Besides, leveraging the impressive power of language models~\cite{chatgpt,palme,saycan,3dllm,bert} in high-level task planning, our method leads to a comprehensive embodied application.

Existing works for mobile manipulation include neural policies~\cite{alfred,mtrack,et,rt1} and map-based planning~\cite{hlsm,film,prompter}.
The former requires numerous training trajectories and annotations of high costs and suffers from the conflict of long-horizon nature and memory-less perception.
The latter lacks flexibility in execution and hardly self-adapts in running time.
To this end, we present \textbf{DISCO} (\textbf{DI}fferentiable Scene \textbf{S}emantics and Dual-level \textbf{CO}ntrol).
It learns dynamic scene representations of objects and affordances on the fly, which facilitates map-based coarse navigation planning.
A neural policy is deployed next to perform fine controls and boost object interaction.

Learning a dependable spatial representation of the scene is crucial for robotic applications, which can be 2D top-down view~\cite{film} or 3D voxels~\cite{hlsm} in practice. 
An ideal scene representation should embody the following attributes: 
(1) Rich semantics in objects and affordances. It encapsulates objects and potential actions in the spatial space. 
(2) On-the-fly update. The scene is always dynamic and subject to changes upon interaction. 
(3) Easy accessibility for queries. It can be queried to facilitate downstream tasks like map-based trajectory planning. 
(4) Generalizability. The representation can be learned in unseen scenes.
To the best of our knowledge, previous research in mobile robotics has hardly developed representations encompassing all these attributes.
In contrast, we build differentiable scene representations with all these attributes and demonstrate its prowess in the realm of interactive navigation.

Recent large models~\cite{rt1,rt2} equipped with scaled datasets and model capacity, have demonstrated successes in generating end-to-end neural actions.
However, when faced with limited data, neural policies~\cite{et,mtrack} struggle on academic benchmarks~\cite{alfred}, primarily due to the data-hungry issue.
Besides, mobile manipulation tasks often involve lengthy trajectories, yet semantic objectives within egocentric observations are rare. 
This scarcity presents significant challenges to neural policies.
To address this, we direct actions based on global and local spatial cues and formulate dual-level coarse-to-fine controls.
First, we design analytical controls on the map to drive the agent coarsely toward the object, with global cues from the scene representations.
Subsequently, a fine-grained short-horizon neural control tailored to local egocentric observation is designed to fine-tune the pose and manipulate the object efficiently.
Prior works use manually-defined rules~\cite{film} or human-in-the-loop feedback~\cite{lgs} to perform some sort of adjustments. But they are hard to formulate and scale up.
Our dual-level control paradigm reduces the need for lengthy action trajectories and enhances overall efficiency.

DISCO can easily integrate into embodied applications such as embodied instruction following, where an embodied agent takes multimodal inputs to accomplish mobile manipulation tasks.
We deploy DISCO on the widely-used ALFRED~\cite{alfred} benchmark consisting of long-horizon vision-language navigation and interaction tasks simulated in AI2THOR~\cite{ai2thor} environment as a testbed.  
ALFRED incorporates high-level human language directives to define the ultimate goal of each task, coupled with low-level step-by-step instructions for agent planning, as depicted in \cref{fig:teaser}.
In this work, we also challenge the planning ability in long-horizon tasks under a setting that omits low-level instructions.
In extensive experiments, we achieve a substantial 11.0\% gain of success rate in unseen scenes without step-by-step instructions, which even outperforms the state-of-the-art method that uses step-by-step instructions by 8.6\%.
We have also made thorough analyses and qualitative studies for a comprehensive evaluation.

Our contributions include: 
\textbf{(1)} We develop differentiable representations enriched with object and affordance semantics. It is dynamic, easy to query, and can be easily deployed in unseen environments.
\textbf{(2)} We propose a dual-level approach that integrates both global and local cues for coarse-to-fine controls, enabling efficient mobile manipulation within limited imitation data.
\textbf{(3)} In extensive experiments, we have evaluated our agent on ALFRED~\cite{alfred} benchmark and achieved new \textit{state-of-the-art} performance with sizeable improvements.

\section{Related Works}
\label{sec:related_work}

\noindent\textbf{Embodied Navigation and Interaction.}
Mobile robotics requires fundamental ability in navigation and interaction. 
In past years, many simulators, scene assets, and benchmarks ~\cite{ai2thor,alfred,habitat19iccv,igibson2,rxr,r2r,RoomR,touchdown,iqa,eqa,zhu2017visual,chen2020soundspaces,ehsani2023imitating,homerobot,behavior}, including both indoor and outdoor scenes, have been developed to facilitate algorithm researches in embodied AI. 
Early works only require navigation in static environments, such as PointNav~\cite{habitat2020sim2real} and ObjectNav~\cite{batra2020objectnav}. 
However, subsequent studies~\cite{igibson2,RoomR,alfred} have expanded to interactive navigation, with semantic changes in dynamic scenes and object manipulation.
The body of related works can be diversified into different categories based on modalities.
Some employ natural language as instructions, as seen in~\cite{r2r,rxr,alfred}.
In contrast, other works like SoundSpace~\cite{chen2020soundspaces} utilize audio instructions, and DialFRED~\cite{DialFRED} incorporates dialogue into navigation. 
Besides, embodied mobile manipulation agents should possess the ability to perform complex reasoning tasks, such as room rearrangement~\cite{RoomR} and question answering~\cite{iqa,eqa,alfworld} as well.
The long-term target of indoor navigation and interaction research is to build agents to accomplish everyday household activities like humans~\cite{behavior}.

\noindent\textbf{Works on ALFRED.}
As a widely-used interactive vision-language navigation benchmark, ALFRED~\cite{alfred} attracts much interest from the research community.
Existing works on ALFRED can be divided into model-free~\cite{alfred,abp,moca,et,lwit,hitut,mtrack} and model-based~\cite{hlsm,film,epa,lgs,prompter} schools. 
The former school deploys an end-to-end neural policy to generate actions, necessitating extensive and costly training trajectories and instructional annotations.
Early works~\cite{alfred,et} utilized LSTM or Transformer to encode visual observation, language instructions, and action history for next-step prediction. They are trained to mimic expert behavior but exhibit subpar performance in novel environments. Some other techniques like panoramic observation~\cite{abp} and instruction alignment~\cite{mtrack} have been proposed for improvement.
In contrast, the latter school constructs a model of the scene to facilitate action planning.
The modeled scene can be a representation of 3D voxels~\cite{hlsm} or 2D top-down views~\cite{film,lgs,prompter}. 
Object search modules~\cite{film,prompter} were designed to help agents find objects.
In-context planning and memory were explored in~\cite{kim2023context}.
In our work, the novel dual-level control utilizes the advantages of both sides.

\noindent\textbf{Affordance.}
Affordance~\cite{gibson} reveals the potential interactions in the physical world. 
It is a multidisciplinary concept of vision, cognition, and robotics. 
Affordance can be learned from 2D~\cite{ocl} and 3D~\cite{partafford,deng20213d} visual contents, language model reasoning~\cite{voxposer}, experienced interactions~\cite{nagarajan2020learning} and reinforced value estimations~\cite{saycan}.
This concept finds a variety of robotic applications, such as scene exploration ~\cite{nagarajan2020learning}, optimal view selection~\cite{li2023imagemanip}, and mobile manipulation~\cite{rt1,adaafford,voxposer}.
In our study, we utilize the ground-truth knowledge from the embodied simulator and learn affordances through supervised learning.

\noindent\textbf{LLMs for Mobile Manipulation.}
The recent advancements in Large Language Models (LLMs) suggest their substantial potential in scaling robotic mobile manipulations.
LLMs contribute significantly to mobile agents by facilitating scene understanding~\cite{3dllm}, task planning~\cite{tapa,palme}, affordance grounding~\cite{saycan} and decision making~\cite{rt1,rt2}.
They can be integrated into the robotic navigation framework via prompted query~\cite{navgpt}, multi-expert discussion~\cite{long2023discuss}, or fine-tuning~\cite{tapa}, paving for more comprehensive robots.

\begin{figure*}[t]
\centering
\includegraphics[width=0.95\textwidth]{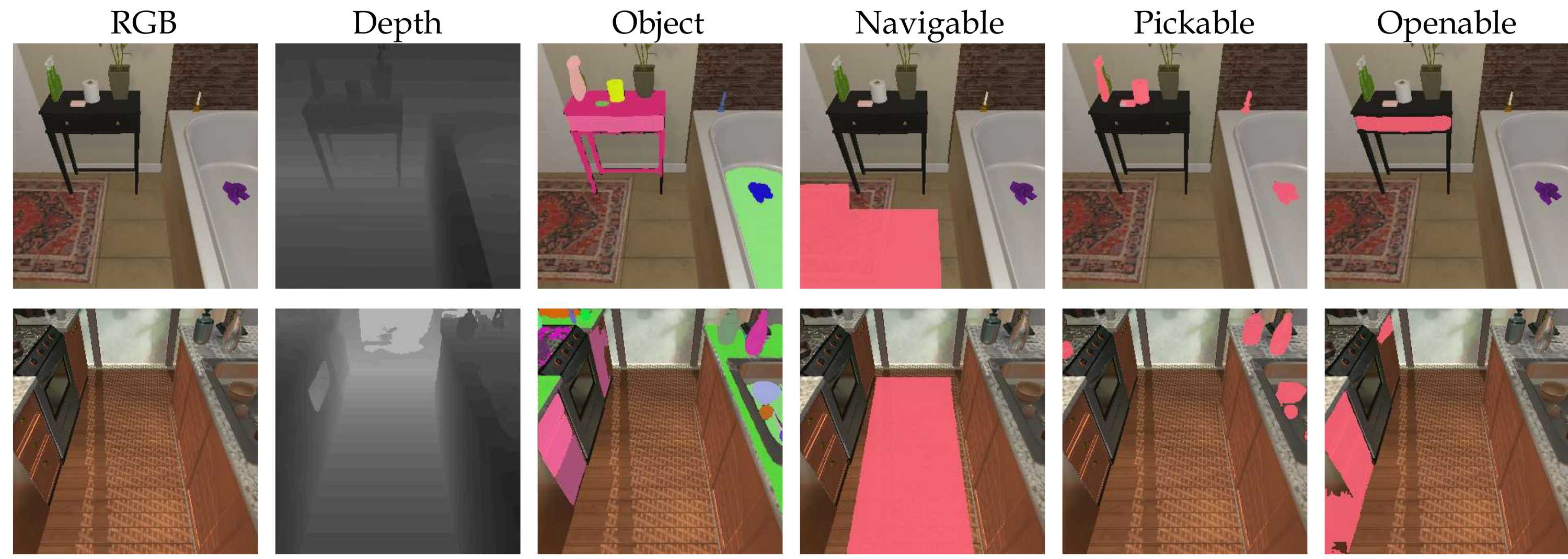}
\caption{
\textbf{The perception foundation.} 
\textbf{(i) 1\textsuperscript{st} column:} egocentric RGB frames as the initial observation.
\textbf{(ii) 2\textsuperscript{nd} column:} depth estimations.
\textbf{(iii) 3\textsuperscript{rd} column:} object instance segmentations.
\textbf{(iv) 4\textsuperscript{th}-6\textsuperscript{th} columns:} affordance masks predictions: the navigable mask and two interactable masks (namely pickable and openable) as references.
}
\label{fig:perception}
\end{figure*}
\section{Approach}
\label{sec:approach}
We introduce how DISCO works in this section, including the perception system in \cref{sec:perception}, the scene representation in \cref{sec:scene}, the dual-level coarse-to-fine controls in  \cref{sec:control}, and its application to Embodied Instruction Following in \cref{sec:eif}.

\begin{figure*}[h]
    \centering
    \includegraphics[width=.98\textwidth]{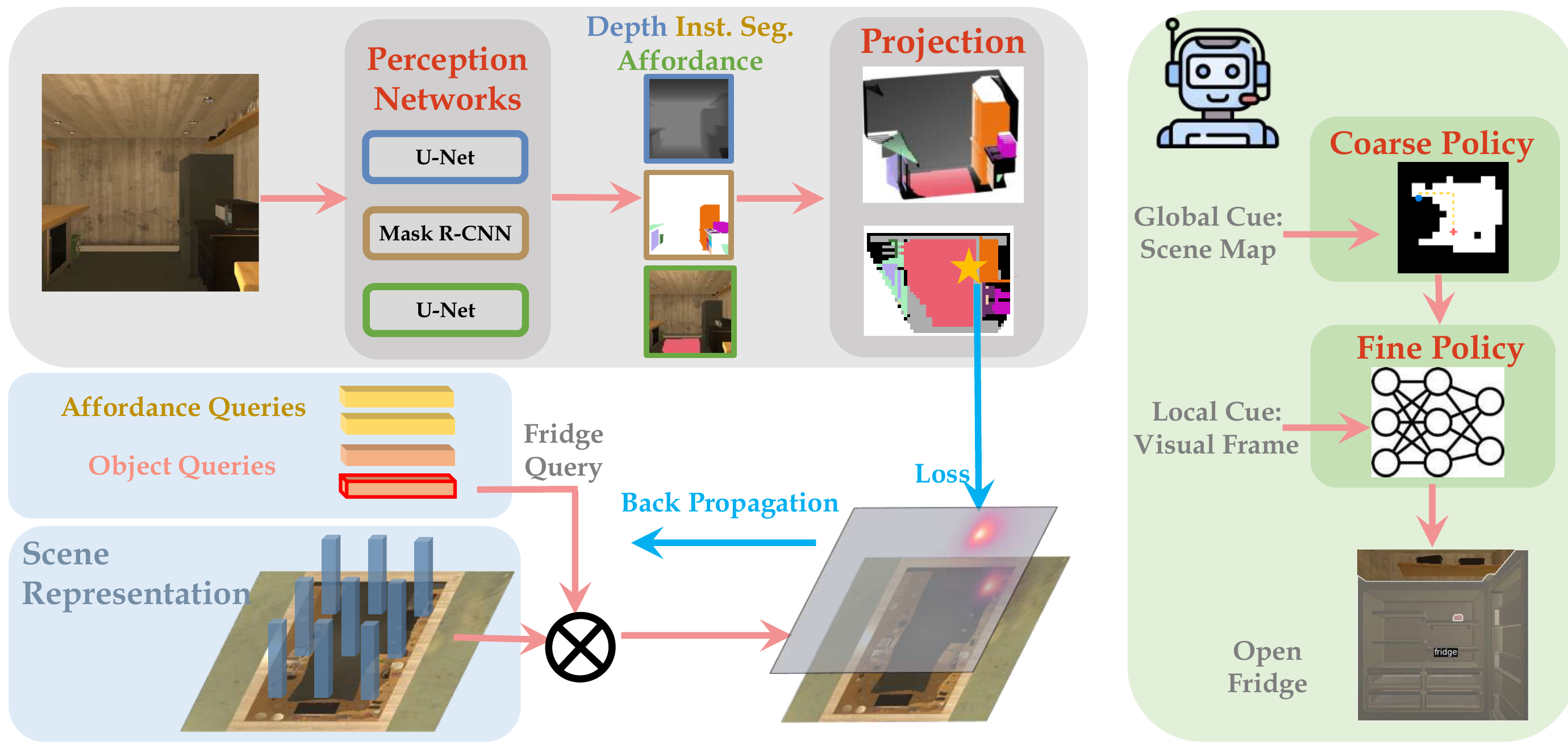}
    \caption{
    \textbf{An overview of the DISCO framework.} 
    Starting from the egocentric RGB frame, our perception system predicts pixel-wise depth, instance segmentation, and affordance frames. 
    They are converted into semantic point clouds via projection and localized in the scene. 
    We build differentiable scene representations with semantic queries to model the scene. They are optimized using gradient descent to match localized point cloud semantics.
    We apply dual-level coarse-to-fine controls.
    The coarse control depends on the global semantic map to approach the localized target. The fine control leverages a neural policy based on the local visual frame to interact.
    }
    \label{fig:overview}
\end{figure*}

\subsection{Perception}
\label{sec:perception}
Our agent perceives the surroundings from an egocentric RGB frame. 
We use three neural nets to estimate finer-grained spatial information: depth, instance segmentation, and affordances.
All three generate pixel-wise information about the frame.
\cref{fig:perception} gives an intuitive example.
We deploy a Mask R-CNN~\cite{maskrcnn} to detect objects, and two U-Nets~\cite{unet} to estimate depth and affordances respectively. The architecture settings of segmentation and depth estimation modules are the same as baselines~\cite{film,hlsm, prompter}.

Training reliable perception modules requires a substantial amount of high-quality curated data. 
They are of significant cost in the real world, but in our study, we collect data via querying a simulation oracle in AI2THOR~\cite{ai2thor}. 
The ground-truth depth is from the depth sensor. 
The ground-truth object segmentation of 85 classes is from the projection of the simulated objects in AI2THOR.
The collected affordance includes one navigation class and seven interaction classes. 
For navigability, we discretize the scene into grids of $25 cm\times 25 cm$, aligning with the agent moving step size. Then we teleport our agent to traverse over all grids and determine navigable ones. Pixels in the egocentric view that are localized in the navigable grids form the navigability mask.
For interactivity, we directly query actionable properties of objects (\eg \textit{Openable}) from AI2THOR and merge all actionable objects in the view into the affordance mask of a certain interaction class.

We collect all frames from training trajectories to train perception networks, while unseen scenes are strictly prohibited. 
The Mask R-CNN~\cite{maskrcnn} network for instance segmentation is initialized from a COCO~\cite{coco} pre-trained checkpoint and then is finetuned by AdamW for 15 epochs with base learning rate 2e-4 and batch size 60. Default Mask R-CNN losses are adopted. A linear warm-up~\cite{loshchilov2017sgdr} of the learning rate is used in the first 1,000 steps.
The U-Net~\cite{unet} for depth estimation is optimized by AdamW for 15 epochs with base learning rate 1e-3 and batch size 80. The depth of each pixel is discretized into 50 bins of $10 cm$ each and trained via a cross-entropy loss.
The U-Net~\cite{unet} for affordance estimation is optimized by AdamW for 25 epochs with base learning rate 1e-3 and batch size 80. We use binary cross-entropy loss to supervise all classes.

\subsection{Learning Scene Representations}
\label{sec:scene}

Prior works~\cite{hlsm,film,prompter} mainly utilize cell-based representations to model the scene. However, discrete cells suffer from imperfect perception issues, \eg hand-crafted rules are required to fix an object miss in one frame. 
We leverage continuous features to learn more robust scene representations. It softly models the scene map with a trade-off between historical and current observation. Different from the matching mechanism in previous continuous representation work~\cite{csr}, we use gradients to update the scene.

We model the scene as a $20 m \times 20 m$ room and discretize it into $25 cm \times 25 cm$ squares, which leads to $M \times M (M=80)$ grids in total. This configuration can cover all scenes in AI2THOR~\cite{ai2thor} and aligns the moving step size.
Each grid is allocated a $C-$dimensional($C=256$ in our implementation) embedding. 
Additionally, we initialize $N^o + N^a$ semantic queries of $C$ dimensions each, where $N^o$ is the number of object classes and $N^a$ is the number of affordance classes.

Our scene representation facilitates straightforward querying and yields a semantic map.
Let $s_{i} (i=1,2\cdots,M^2)$ be the scene representation of the $i$-th grid, $q_j (j=1,2\cdots,N^o+N^a)$ be the $j$-th semantic query vector. 
We obtain the probability $p_{i,j}$ of the $j$-th class at $i$-th grid by a composition  function $f$, such as $p_{i,j} = f(s_i, q_j)$.
For system efficiency, we employ a minimal query mechanism via inner-dot followed by sigmoid, \ie $f(s_i, q_j) = \sigma (s_i^T~q_j)$, where $\sigma$ is the sigmoid function.
We apply zero-value initialization for $s_i$ while random initialization from a normal distribution for $q_j$, thus it predicts $p_{i,j}=0.5$ of no certainty at the beginning of the episode.

A scene is usually dynamically changed subject to embodied interactions.
Therefore, a key factor of our scene representation is on-the-fly optimization at each step. 
We illustrate the learning framework in \cref{fig:overview} and describe the more detailed process below.
Starting from the egocentric RGB frame, our perception system predicts extra depth, segmentation, and affordance frames of rich geometric and semantic information. 
Following the semantic mapping procedure in~\cite{chaplot2020learning,film}, the egocentric frame is converted to point clouds with semantics via camera projection. 
Notably, AI2THOR uses a discrete action space thus we estimate camera pose by the accumulation of historic actions.
Top-down projection is followed to generate an allocentric point map of the current observation. 
The localized semantic-aware points are used to supervise the scene optimization.
Let $c_i$ be localized points in $i$-th grid and $c_i^j$ be the points of $j$-th semantics among them. 
Then $c_i^j / c_i$ is the proportion of semantic points. 
We normalize the proportion to get the soft grid-level semantic label $y_i^j$:
\begin{equation}
    y_i^j = \frac{c_i^j / c_i}{\max_{1\le k\le M^2} \{c_k^j / c_k\}},
    \label{eq:label}
\end{equation}
where the positive label $1$ is assigned for the largest semantic proportion and other labels decrease. 
We optimize representations of visible grids at each step while leaving other grids unchanged. We threshold localized point cloud density to determine visibility, \ie the visibility of $i$-th grid $v_i=\mathbf{1}[c_i > \rho]$ where $\mathbf{1}[\cdot]$ is the binary indicator function and $\rho$ ($\rho=500$ in our implementation) is the threshold.
With all the estimated utilities above, we optimize the scene using back-propagation. 
For visible grids with $v_i=1$, we compute the cross-entropy loss between $y_i^j$ and $f(s_i,q_j)$ to update the feature. The learning step is formulated as:
\begin{gather}
L(y_i^j, f(s_i,q_j)) = -(1-y_i^j)(1-f(s_i,q_j)) - y_i^j f(s_i,q_j),\\
s_{i} \longleftarrow s_i - \alpha \cdot \sum_j v_i \cdot \frac{\partial}{\partial s_i}  L(y_i^j, f(s_i,q_j)),\\
q_{j} \longleftarrow q_j - \alpha \cdot \sum_i v_i \cdot \frac{\partial}{\partial q_j}  L(y_i^j, f(s_i,q_j)).
\end{gather}
For each step, we update $s_i$ and $q_j$ for 10 learning iterations with learning rate $\alpha=0.01$.

Till now, we have built scene representations that can be optimized by semantic scene differentiation. It omits some hand-crafted scene-updating strategies in prior work~\cite{film} and proves to be more generalizable.

\begin{figure}[t]
    \centering
    \includegraphics[width = 0.75\linewidth]{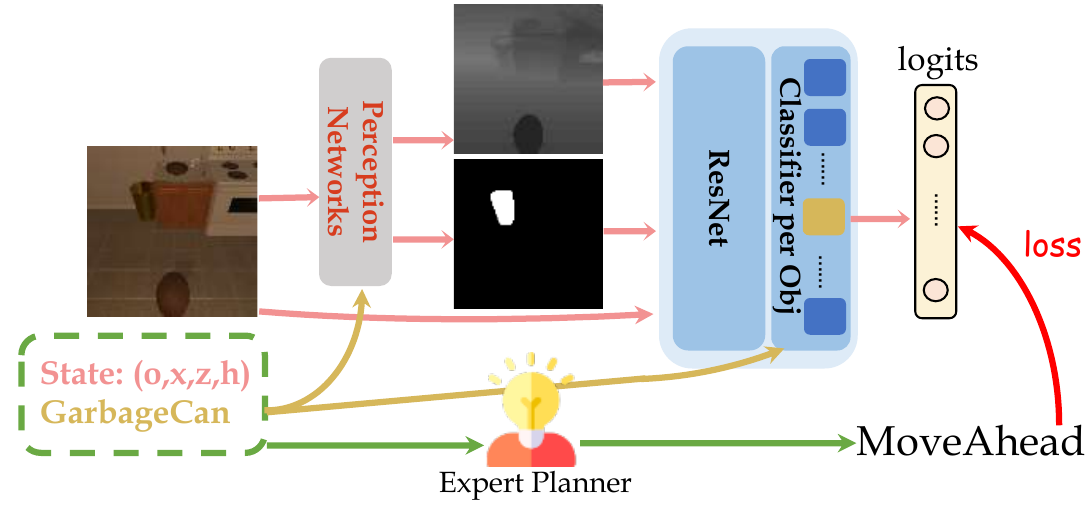}
    \caption{The design of fine action control. DISCO employs a neural policy network to predict fine action steps. RGB, depth, and the object mask are sent to the network to derive a feature, followed by an object-specific classifier to predict the action. The policy is trained by mimicking expert actions.}
    \label{fig:policy}
\end{figure}

\subsection{Coarse-to-Fine Action Control}
\label{sec:control}
In this section, we introduce how DISCO acts to accomplish a primitive task represented as a verb-noun pair, \eg \textit{Pickup Lettuce}. 
It first starts with random walks until the target object is observed. Then we execute coarse-to-fine controls to perform mobile manipulation, where the coarse control navigates to approach near the object and fine control refines the agent pose for interaction.

\noindent\textbf{$\bullet$ Random Walk.} 
Our agent begins with a random walk if the object has never been detected by the detector. 
We use the navigable query (one of the affordance classes) to obtain the navigability map of the scene. 
Then, we randomly select a navigable waypoint and apply the Breadth First Search (BFS) algorithm to plan a trajectory to the destination.

\noindent\textbf{$\bullet$ Coarse Control.}
The random walk terminates once the object is spotted by the detector. Then, we apply map-based coarse control to navigate the agent close to the localized target. 
The coarse control is based on global cues, \ie semantic scene maps.
We query the scene representations to obtain probabilistic maps of the semantic objects and affordances of all grids, where affordances play a supplement role in localizing a desired interaction.
Next, we select the grid of the largest object-affordance union probability as the target. 
For instance, when the agent is asked to \textit{Pickup Lettuce}, it queries the object \textit{Lettuce} distribution in the scene as well as the affordance \textit{Pickupable} distribution, then the grid of maximal multiplied union probability is targeted. 
Our coarse control is designed to approach the object. 
We expand the target location and its neighboring grids within \textit{1m} distance to be the destination of coarse navigation.
BFS algorithm based on the navigability map is applied to plan trajectories.

\noindent\textbf{$\bullet$ Fine Control.}
Though map-based coarse actions are efficient in planning lengthy navigation trajectories.
It can't be self-adapted to manipulate objects.
The local state, such as view direction and distance to the object, greatly affects whether an interaction can be successful.
Some refinement based on local cues, \ie egocentric frames, are essential for better manipulations.
To this end, we propose fine action controls by a neural network, illustrated in \cref{fig:policy}.  
In the fine action step, we formulate each state as $(o,x,z,h)$, where $o$ is the target object going to be manipulated, $(x,z)$ is the agent location and $h$ is the camera horizon.
We adjust the agent direction to the target object by referring to the localized yaw first.
This makes the target object visible.
We use the concatenation of RGB, estimated depth, and the mask of the target object to be the input of the policy. The input includes geometric distance and identifies the target object.
We use a ResNet50~\cite{resnet} to encode the feature, followed by object class-specific classifiers to generate actions.
The independent classifier design is motivated by the widely used object detector head design~\cite{maskrcnn}.
We find some insights into neural policy learning. First, we adjust agent rotation to keep the object in view, which reduces the ambiguity of control. Second, our neural policy is applied exclusively in short-horizon refinement stages, simplifying the learning process for lengthy trajectories.
Notably, existing works~\cite{film,prompter} use hand-crafted rules or tests to deal with the openness of openable receptacles. Our affordance module (\cref{fig:perception}) automatically detects openable property and helps DISCO to act.

We build an expert planner with full knowledge of all scenes to help policy learning by imitation. 
The expert building process follows the generative pipeline of ALFRED~\cite{alfred}.
It can label all interactable states and plan transiting actions by BFS search.
Since coarse control navigates the agent close to the target within $1m$, the fine control next only requires short-horizon refining steps, which eases the difficulty of training.
We collect data on states within 4 expert steps to interactions.
We iterate over all objects and all short-horizon states to save frames, generating a training set of 316,935 frames. 
As default training trajectories of ALFRED~\cite{alfred} contain 1,051,308 frames, we find training short-horizon policy is more data-efficient.
We train the policy by imitating planned actions from the expert, also known as behavior cloning. 
It is supervised by expert actions using a cross-entropy loss. 
We optimize the policy using an AdamW optimizer with a constant learning rate 5e-5 and train it for 40 epochs with a batch size of 100.

\subsection{Application: Embodied Instruction Following}
\label{sec:eif}

DISCO performs primitives of mobile manipulation tasks commanded by verb-noun pairs and can be easily applied in diverse embodied tasks.
We take the embodied instruction following tasks from ALFRED~\cite{alfred} as a test bed, where the agent is commanded by natural language instructions to accomplish long-horizon tasks of many mobile manipulation subgoals.
More introduction about the benchmark can be found in the supplementary material.
However, though we take vision-language navigation and interaction to conduct main experiments, we believe DISCO has the potential for other modality instructions like sound~\cite{chen2020soundspaces} or dialog~\cite{DialFRED} with different instruction processing modules.

In the detailed implementation, for a fair comparison with the baseline, we inherit the instruction processing procedure and some building components from FILM~\cite{film}.
Notably, though ALFRED provides low-level step-by-step instructions, our methods can also run without these annotations.
High-level goal directives are only used by default to generate subgoal plans.
To generate plans, we adopt fine-tuned BERTs~\cite{bert} from~\cite{film} to parse natural language instructions into ALFRED internal parameters. 
Next, leveraging the patterned task nature of ALFRED, templates are used to convert parameters into multiple verb-noun manipulation subgoals.
For instance, we take the case in \cref{fig:teaser} as an example. 
Language models recognize it as a \textit{pick\_clean\_then\_put} task with object argument \textit{Lettuce} and receptacle argument \textit{DinningTable}. 
It is converted to subgoal series: (\textit{Pick}, \textit{Lettuce}), (\textit{Clean}, \textit{Lettuce}), (\textit{Put}, \textit{DinningTable}) using the template.
We give more details about the instruction processing and the holistic application in the supplementary material.


\section{Experiments}
\label{sec:exp}

\subsection{Evaluation Protocols}
We evaluate our method on ALFRED~\cite{alfred}, consisting of large-scale long-horizon vision-language navigation and interaction tasks. 
The dataset is divided into \textit{train}, \textit{validation}, and \textit{test} splits, containing 21,023/1,641/3,062 episodes respectively. 
We use \textit{test} split to compare with competitive baselines, but \textit{valid} split to make analyses. 
Both \textit{valid} and \textit{test} splits are divided into \textit{seen} and \textit{unseen} scenes. 
The \textit{valid/test} splits have 820/1,533 episodes in seen scenes while 821/1528 episodes in unseen scenes.

Four metrics are used for evaluation. 
(1) Success Rate (SR). Rate of accomplished tasks with success. 
(2) Goal Condition (GC). Rate of achieved conditions for goals. 
(3) Path Length Weighted SR (PLWSR). Weighing SR by the agent trajectory length against expert trajectory length. 
(4) Path Length Weighted GC (PLWGC). Applying the same weight on GC. 
SR and GC mainly reflect the effectiveness of agents while PLW metrics reflect the efficiency of running steps. All metrics are the higher the better.

\subsection{Baselines}
We adopt competitive baselines reported on ALFRED to compare with our method. 
They include Seq2Seq~\cite{alfred}, MOCA~\cite{moca}, E.T.~\cite{et}, LWIT~\cite{lwit}, ABP~\cite{abp}, M-Track~\cite{mtrack}, HiTUT~\cite{hitut}, HLSM~\cite{hlsm}, FILM~\cite{film}, LGS-RPA~\cite{lgs}, EPA~\cite{epa}, prompter~\cite{prompter}, and CAPEAM~\cite{kim2023context}. 
All methods use RGB vision and language instructions in test time.
While step-by-step instructions can be omitted by some methods as well as DISCO, we separate the comparison for fairness.
By default, DISCO only uses a high-level description of the goal.

\begin{table}[!t]
\caption{Results in the test splits of ALFRED~\cite{alfred}.}
\begin{center}
\resizebox{0.93 \linewidth}{!}{
  \begin{tabular}{lc|cccc|cccc}
    \toprule
    &  step-by-step     &\multicolumn{4}{c|}{\bf Test Seen} &\multicolumn{4}{c}{\bf Test Unseen}\\    
    & instructions      & SR & GC & PLWSR & PLWGC & SR & GC & PLWSR & PLWGC\\
    \midrule
    Seq2Seq~\cite{alfred}       &\CheckmarkBold   & 4.0 & 9.4 & 2.0 & 6.3 & 3.9 & 7.0 & 0.1 & 4.3\\
    MOCA~\cite{moca}            &\CheckmarkBold   & 26.8 & 33.2 & 19.5 & 26.8 & 7.7 & 15.7 & 4.2 & 11.0\\
    E.T.~\cite{et}              &\CheckmarkBold   & 38.4 & 45.4 & 27.9 & 34.9 & 8.6 & 18.6 & 4.1 & 11.5\\
    LWIT~\cite{lwit}            &\CheckmarkBold   & 29.2 & 38.8 & 24.7 & 34.9 & 8.4 & 19.1 & 5.1 & 14.8\\
    HiTUT~\cite{hitut}          &\CheckmarkBold   & 21.3 & 30.0 & 11.1 & 17.4 & 13.9 & 20.3 & 5.9 & 11.5\\
    ABP~\cite{abp}              &\CheckmarkBold   & 44.6 & 51.1 & 3.9 & 4.9 & 15.4 & 24.8 & 1.1 & 2.2 \\
    FILM~\cite{film}            &\CheckmarkBold   & 27.7 & 38.5 & 11.2 & 15.1 & 26.5 & 36.4 & 10.6 & 14.3\\
    M-Track~\cite{mtrack}       &\CheckmarkBold   & 24.8 & 33.3 & 13.9 & 19.5 & 16.3 & 22.6 & 7.7 & 13.2\\
    LGS-RPA~\cite{lgs}          &\CheckmarkBold   & 40.1 & 48.7 & 21.3 & 29.0 & 35.4 & 45.2 & 15.7 & 22.8\\
    Prompter~\cite{prompter}    &\CheckmarkBold   & 53.2 & 63.4 & 25.8 & 30.7 & 45.7 & 58.8 & 20.8 & 26.2\\
    CAPEAM~\cite{kim2023context}&\CheckmarkBold   & 51.8 & 60.5 & 21.6 & 25.9 & 46.1 & 57.3 & 19.5 & 24.1\\
    {\bf DISCO (Ours)}          &\CheckmarkBold   & \textbf{59.5} & \textbf{66.1} & \textbf{40.6} & \textbf{47.4} & \textbf{56.5} & \textbf{66.8} & \textbf{36.5} & \textbf{44.5}  \\
    \midrule
    HiTUT~\cite{hitut}          &\XSolidBrush     & 13.6 & 21.1 & 5.6 & 11.0 & 11.1 & 17.9 & 4.5 & 9.8\\
    HLSM~\cite{hlsm}            &\XSolidBrush     & 25.1 & 35.8 & 6.7 & 11.5 & 16.3 & 27.2 & 4.3 & 8.5\\
    FILM~\cite{film}            &\XSolidBrush     & 25.8 & 36.2 & 10.4 & 14.2 & 24.5 & 34.8 & 9.7 & 13.1\\
    LGS-RPA~\cite{lgs}          &\XSolidBrush     & 33.0 & 41.7 & 16.7 & 24.5 & 27.8 & 38.6 & 12.9 & 20.0\\
    EPA~\cite{epa}              &\XSolidBrush     & 40.0 & 44.1 & 2.6 & 3.5 & 36.1 & 39.6 & 2.9 & 3.9\\
    Prompter~\cite{prompter}    &\XSolidBrush     & 49.4 & 55.9 & 23.5 & 29.1 & 42.6 & 59.6 & 19.5 & 25.0\\
    CAPEAM~\cite{kim2023context}&\XSolidBrush     & 47.4 & 54.4 & 19.0 & 23.8 & 43.7 & 54.6 & 17.6 & 22.8\\
    {\bf DISCO (Ours)}          &\XSolidBrush     & {\bf 58.0} & {\bf 64.9} & {\bf 39.6} & {\bf 46.5} & {\bf 54.7} & {\bf 65.5} & {\bf 35.5} & {\bf 43.6}\\
    \bottomrule
  \end{tabular}

}
\end{center}
\label{tab:result}
\end{table}

\subsection{Quantitative Comparisons}
We report the quantitative results of DISCO and competitive baselines in test splits of ALFRED in  \cref{tab:result} for fair comparisons. 
We find our method outperforms the art by sizable improvements in all slots. 
Equipped with step-by-step instructions, our method achieves 59.5\% and 56.5\% success rates in seen and unseen scenes, with substantial gains of 6.3\% and 10.4\% over the state-of-the-art methods, showcasing the effectiveness of DISCO. 
The superiority is consistent with other metrics. 
Integrating the path length into metrics, we achieve almost 1.57x and 1.75x performances than Prompter~\cite{prompter} on PLWSR metrics in seen and unseen scenes, which means our agent accomplishes tasks using fewer steps in execution. This strongly validates the efficiency.
Besides, in terms of goal condition metrics, we also have superior performances over baselines, namely 66.1\% and 66.8\% in seen and unseen scenes respectively.
Next, we omit step-by-step instructions and challenge DISCO only using high-level goals.
Under this setting, our method achieves 58.0\% and 54.7\% success rates in seen and unseen scenes, outperforming baselines by 8.6\% and 11.0\%. 
A bigger surprise is that DISCO without step-by-step instructions outperforms the art with step-by-step instructions, where we achieve success rate gains of 4.8\% and 8.6\% in seen and unseen scenes respectively.
It proves we surpass all existing methods with non-trivial advancement in both efficiency and effectiveness.
Overall, our method establishes new \textit{state-of-the-art} performances on ALFRED.

\begin{table}[t]
\centering
\begin{minipage}[t]{0.6\textwidth}
    \centering
    \caption{Ablation study.}
    \vspace{-0.3cm}
    \label{tab:abl}
    \resizebox{\textwidth}{!}{
          \begin{tabular}{l|cc|cc}
    \toprule
         &\multicolumn{2}{c|}{\bf Valid Seen} &\multicolumn{2}{c}{\bf Valid Unseen}\\    
     & SR & GC & SR & GC \\   
     \midrule
    DISCO                    & 57.3 & 63.9 & 55.0 & 65.5 \\ 
     \midrule
    + step-by-step instr.    & 65.1 & 70.8 & 59.1 & 68.6 \\
    + gt. lang.              & 70.5 & 75.5 & 64.1 & 71.9 \\
    + gt. percep.            & 67.2 & 73.4 & 66.8 & 73.9 \\
    + gt. percep. lang.      & 79.9 & 84.2 & 79.6 & 83.0 \\ 
     \midrule
    \textit{w.o.} differentiable         & 47.4 & 54.0 & 42.7 & 52.3 \\
    \textit{w.o.} interactive aff.       & 52.5 & 57.9 & 51.3 & 56.6\\
    \textit{w.o.} navigation aff.        & 48.1 & 56.1 & 46.0 & 54.8 \\
    \textit{w.o.} coarse control         & 13.5 & 16.2 & 9.1  & 10.3 \\
    \textit{w.o.} fine control           & 53.0 & 58.3 & 51.7 & 57.3 \\
    \bottomrule
  \end{tabular}

    }
\end{minipage}
\end{table}

\subsection{Ablation Study}
We conduct comprehensive ablation studies on the valid splits of ALFRED to explore the effects of components of DISCO.
Results are reported in \cref{tab:abl}. 

\noindent\textbf{Increments of stronger multimodal inputs.}
We augment DISCO with stronger instruction understanding and perception modules.
Equipped with low-level step-by-step instructions, DISCO achieves 65.1\% and 59.1\% success rates in seen and unseen scenes. Ground-truth language parsing pushes the results to 70.5\% and 64.1\%. This uncovers the potential of stronger language models in boosting embodied planning. 
Ground-truth perceptions further enhance performances to 79.9\% and 79.6\% in seen and unseen scenes.

\noindent\textbf{Differentiable representations.} 
We replace differentiable representations with widely-used cell representations~\cite{film}. 
Noticeable success rate drops of -9.9\% and -12.3\% are observed in seen and unseen scenes. 
This is because our differentiable representations provide soft semantics such as \cref{eq:label} and benefit acting.
In contrast, cell-based representations are binary and require hand-crafted rules to fix inconsistent perceptions.
Our differentiable representations balance historical and current observations to alleviate the issue.

\noindent\textbf{Affordance.} 
We first remove interactive affordances to validate their effects.
In this test, the openable property and affordance-augmented localization are not used.
We achieve 52.5\% and 51.3\% SR in seen and unseen scenes, with drops of -5.1\% and -3.7\%.
Next, we replace the navigability affordance with point-obstacle-based navigation methods.
The performance drops become larger, namely -9.3\% and -9.0\% SR in seen and unseen scenes respectively.
These experiments validate affordance knowledge is crucial for embodied applications.

\noindent\textbf{Dual-level controls.} We remove the dual-level controls and apply coarse or fine control individually. 
Removing coarse control leads to a crash of our system, with extremely low success rates. 
Besides, removing fine control leads to SR performance drops of -4.3\% and -3.3\% in seen and unseen scenes respectively.
The finding suggests that coarse control plays its role in most lengthy moving steps of irreplaceable significance. 
However, the fine control makes some local refinement to facilitate interactions.

\begin{figure}[t]
    \centering
    \includegraphics[width = 0.95\linewidth]{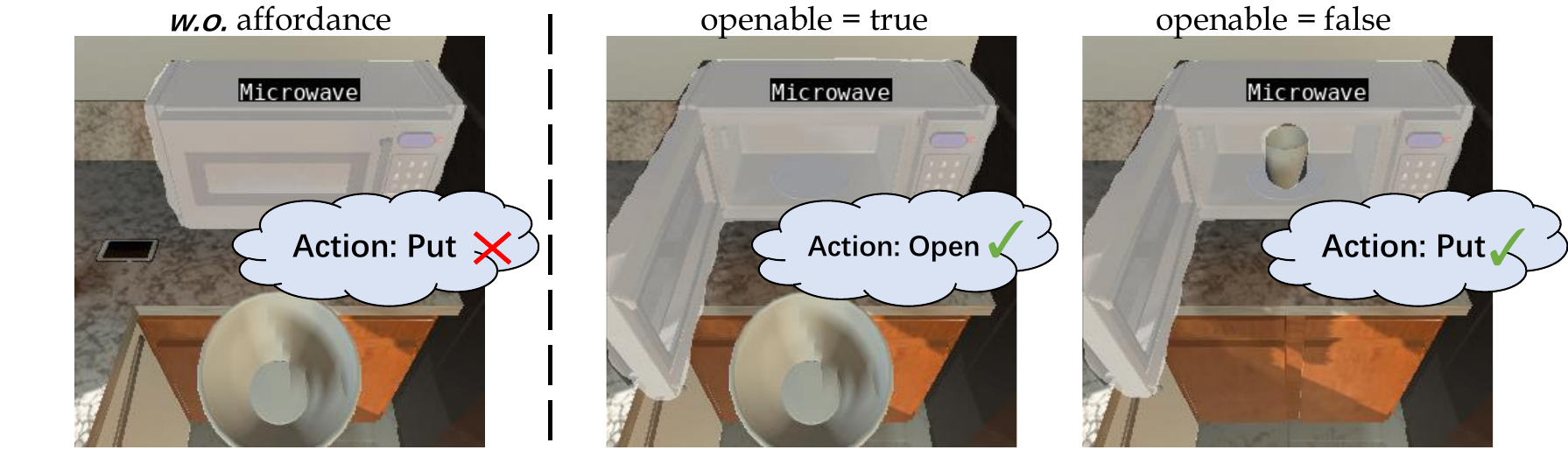}
    \caption{Qualitative case of affordance. Left: The agent fails to put the bowl into the microwave without openable knowledge. Right: DISCO is aware of the openable affordance property in microwave interaction.}
    \label{fig:open}
\end{figure}

\begin{figure}[t]
    \centering
    \includegraphics[width = 0.86\linewidth]{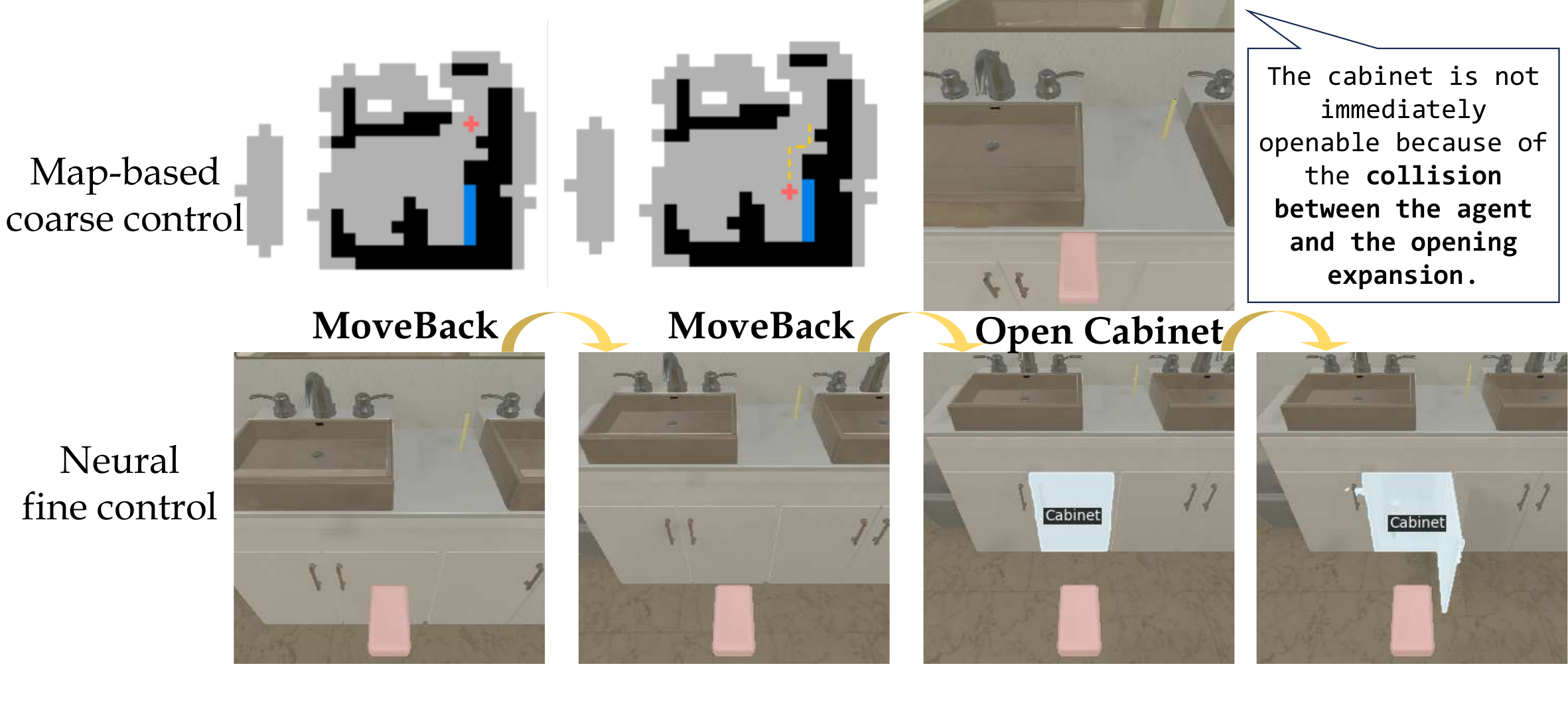}
    \includegraphics[width = 0.86\linewidth]{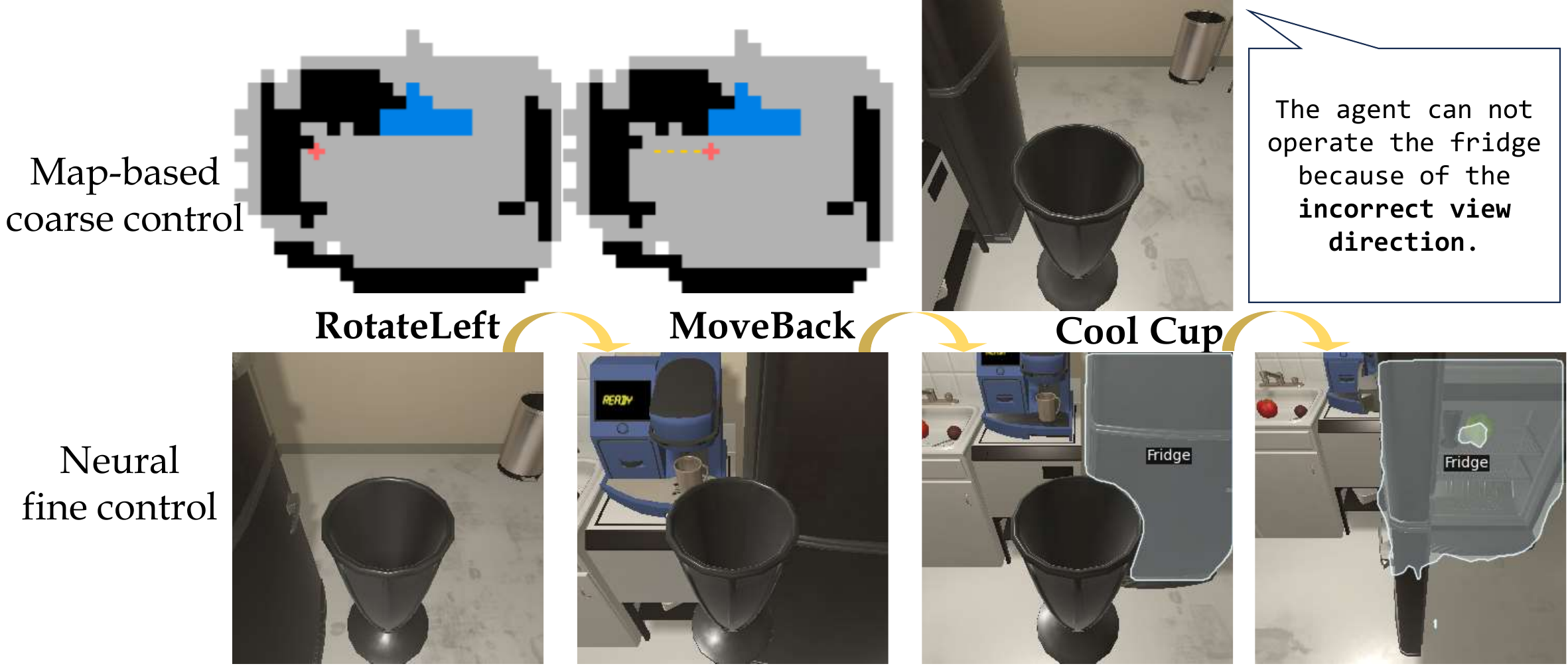}
    \caption{Qualitative running cases. DISCO applies map-based coarse actions followed by neural fine actions. The agent may suffer from the opening collision (upper) or incorrect view direction (bottom) after the coarse navigation. Fine actions perform self-adjustment to facilitate interactions. More cases are in the supplementary material.}
    \label{fig:case}
\end{figure}


\subsection{Qualitative Results}
We provide qualitative cases to reveal some running facts about DISCO.

\noindent\textbf{Affordance.} 
We demonstrate the use of affordance in \cref{fig:open}. 
Prior works use manually defined rules or tests to help interact with openable receptacles. 
Our affordance module can automatically detect the opening state and facilitate acting.
Without the affordance knowledge, the agent cannot place the bowl into the microwave properly.

\noindent\textbf{Dual-level control.}
Our control policy is illustrated in \cref{fig:case}.
DISCO executes map-based coarse actions to approach the object first.
However, the object target is usually not interactable after the coarse stage.
They may be attributed to state change expansion, view direction, position offset, and more diversified reasons.
To this end, we apply neural fine action controls for short-horizon self-adjustment to flexibly address the trouble.
We provide more qualitative results in the supplementary material.

\section{Conclusion}
\label{sec:conclusion}

We introduce \textbf{DISCO} to perform mobile manipulation tasks in this work.
It learns differentiable scene representations of rich semantics in objects and affordances in online exploration.
The scene representations retrieve target semantics and facilitate map-based navigation planning.
We propose dual-level coarse-to-fine action controls, which leverage a global scene map to coarsely approach the navigation target followed by neural fine actions to boost object interaction.
We leverage language models to plan primitive tasks and integrate DISCO into an Embodied Instruction Following application. 
In extensive experiments, DISCO achieves new \textit{state-of-the-art} results on ALFRED.



\section*{Acknowledgments}
This work was supported by the National Key Research and Development Project of China (No. 2022ZD0160102), National Key Research and Development Project of China (No. 2021ZD0110704), Shanghai Artificial Intelligence Laboratory, and XPLORER PRIZE grants.


%
%
\bibliographystyle{splncs04}
\bibliography{main}

\newpage
\appendix
\section{Limitations}
We collect data and conduct experiments in simulation, but more sim2real efforts are necessary in future works. 
We adopt discretized and symbolized action space in AI2THOR. Grounding actions into the physical world with robotic kinematics and dynamics leads to more comprehensive applications. We will continue to work on this path. 

\section{Potential Negative Societal Impacts} 
The proposed embodied AI techniques are not expected to cause significant social harm when utilized appropriately. Nonetheless, the regulations on the safety and privacy aspects of embodied systems remain important.

\section{Introduction of ALFRED}
\label{sec:intro_alfred}
\begin{figure}[t]
    \centering
    \includegraphics[width = 0.99\linewidth]{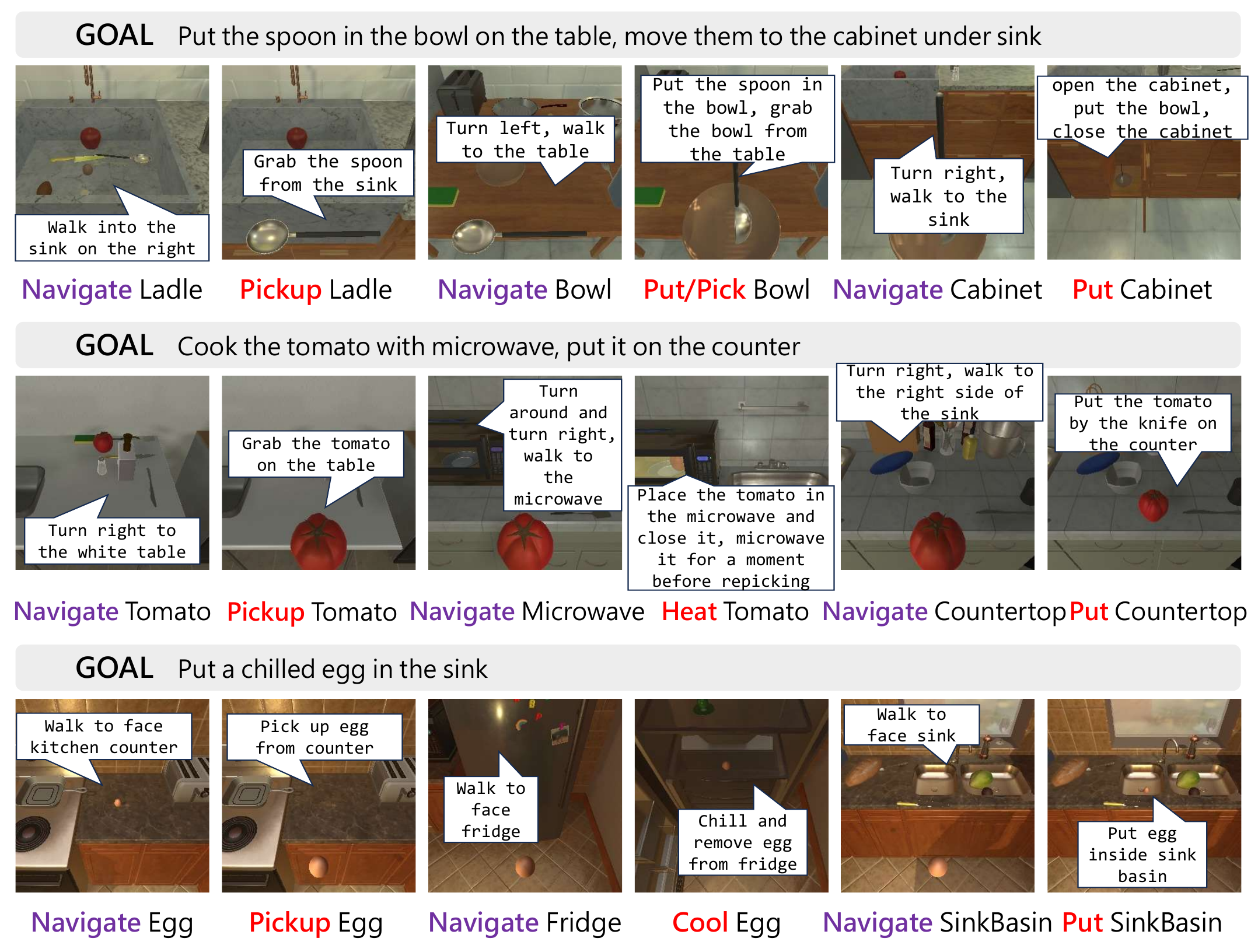}
    \caption{Examples of vision-language navigation and interaction tasks from ALFRED~\cite{alfred}.}
    \label{fig:supp_tasks}
\end{figure}

ALFRED~\cite{alfred} is a large-scale dataset of long-horizon vision-language navigation and interaction tasks.

\noindent\textbf{Task Types.}
ALFRED includes 7 task types:
\begin{itemize}
    \item Look \& Examine. Examine an object under the light (\eg read a book under the light).
    \item Pick \& Place. Pick an object and place it in a receptacle (\eg pick a pencil and place it in a drawer). 
    \item Place Two. Place two object instances in the same receptacle (\eg throw two apples into the garbage bin).
    \item Stack. Place an object in an intermediate container then place the intermediate container in a receptacle (\eg place a tomato in a pot then put the pot on a stove burner). 
    \item Heat \& Place. Place a heated object in a receptacle (\eg place a heated egg on the dining table).
    \item Cool \& Place. Place a cooled object in a receptacle (\eg place a cooled apple on the countertop). 
    \item Clean \& Place. Place a cleaned object in a receptacle (\eg place a cleaned cloth in the bathtub).  
\end{itemize}
We illustrate more task examples in~\cref{fig:supp_tasks}.

\noindent\textbf{Subgoals.}
All tasks mentioned above can be divided into eight subgoals. (1) GotoLocation. (2) PickUp. (3) Put. (4) Slice. (5) Toggle. (6) Heat. (7) Cool. (8) Clean. Each interactive subgoal is an instructed verb-noun pair. The proposed DISCO mainly boosts the execution efficiency of each subgoal as the primitive.

\noindent\textbf{Action Space.}
The action space of ALFRED includes 5 navigation actions ({MoveAhead, RotateRight, RotateLeft, LookUp, LookDown}) and 7 interactive actions (PickUp, Put, Open, Close, ToggleOn, ToggleOff, Slice). We compose three more actions ({MoveLeft, MoveBack, MoveRight}) in the decision space of DISCO:
\begin{itemize}
    \item  MoveLeft = RotateLeft + MoveAhead + RotateRight,
    \item  MoveRight = RotateRight + MoveAhead + RotateLeft,
    \item  MoveBack = RotateLeft + RotateLeft + MoveAhead + RotateRight + RotateRight.
\end{itemize}
Each interactive action requires an object mask to identify the target object to be manipulated. In our framework, the object mask is predicted by Mask R-CNN~\cite{maskrcnn}.

\section{Applying DISCO in ALFRED}
In this section, we describe the holistic process of applying DISCO in ALFRED~\cite{alfred}.

\subsection{Language Processing}
To facilitate a fair comparison, we follow the language processing module in~\cite{film} that employs fine-tuned BERTs~\cite{bert} to map natural language instructions into the internal parameters of ALFRED.

First, it determines one of seven task types, as detailed in~\cref{sec:intro_alfred}, based on the language instruction.
Then it estimates four task arguments:
\begin{itemize}
    \item Object. The object going to be manipulated (\eg apple). 
    \item Receptacle. Where the object will be finally placed (\eg dining table). 
    \item Movable Receptacle. A special parameter to identify the intermediate container in Stack tasks (\eg a pot that holds a tomato and is placed on a stove burner). 
    \item Slice. A boolean parameter to identify whether the object needs to be sliced (\eg slicing lettuce). 
\end{itemize}

The task type and four arguments are the internal parameters for ALFRED tasks, initially defined in the benchmark to form the structured task nature. 
We adopt the off-the-shell fine-tuned BERTs from~\cite{film} to estimate the internal parameters.
Though recent large language models may present potential benefits, our experimental setup ensures a fair comparison with baselines.

\begin{table}[t]
\begin{center}
\caption{Templates on generating subgoals per task type.}
\label{tab:template}
\resizebox{0.45 \linewidth}{!}{
\begin{tabular}{p{1cm}ll}
\toprule
     \multicolumn{3}{l} {\bf (1) Look \& Examine}\\
      & PickUp & Object \\
      & Toggle & Lamp \\
\midrule
     \multicolumn{3}{l} {\bf (2) Pick \& Place}\\
      & PickUp & Object \\
      & Put & Receptacle \\
\midrule
     \multicolumn{3}{l} {\bf (3) Place Two}\\
      & PickUp         & Object \\
      & GotoLocation   & Object \\
      & Put    & Receptacle \\
      & PickUp & Object \\
      & Put    & Receptacle \\
\midrule
     \multicolumn{3}{l} {\bf (4) Stack}\\
      & PickUp         & Object \\
      & Put            & Movable Receptacle \\
      & PickUp         & Movable Receptacle \\
      & Put            & Receptacle \\
\midrule
     \multicolumn{3}{l} {\bf (5) Heat \& Place}\\
      & PickUp & Object \\
      & Heat   & Microwave \\
      & Put    & Receptacle \\
\midrule
     \multicolumn{3}{l} {\bf (6) Cool \& Place}\\
      & PickUp & Object \\
      & Cool   & Fridge \\
      & Put    & Receptacle \\
\midrule
     \multicolumn{3}{l} {\bf (7) Clean \& Place}\\
      & PickUp & Object \\
      & Clean  & Sink Basin \\
      & Put    & Receptacle \\
    \bottomrule
  \end{tabular}
}
\end{center}
\end{table}

\subsection{Task Planning with Templates}

Utilizing the estimated internal parameters from the language module, we capitalize on ALFRED's highly structured task nature to generate subgoals using predefined templates. 
These default templates for each task type are listed in~\cref{tab:template}. 
Each task plan contains multiple primitives in order to achieve the final state.
Each primitive is a verb-noun pair, fed into the proposed DISCO framework in execution.

We note two special points in generating task plans. 
First, in cases where the object is required to be sliced, our approach introduces an initial sequence of three more specific steps: \texttt{(Pickup, Knife)}, \texttt{(Slice, Object)}, and \texttt{(Put, Receptacle)}, positioned at the beginning of the subgoal series. 
Second, for \textit{Place Two} tasks, we insert a \texttt{(GotoLocation, Object)} subgoal to locate the second object instance. 
This strategy is adopted to address the challenge of distinguishing between different instances within the same semantic class. By this approach, the agent is forced to first pick up one object, and then locate the second one before placing the first object down.

\subsection{Agent Setup}

We record the 4-DoF agent pose, comprising a 2-DoF position, a 1-DoF rotation, and a 1-DoF camera horizon degree.
At the beginning of each task episode, we take the initial location of the agent as an anchor point. 
It is the center point of the modeled scene.
We position the camera at a 45-degree downward angle relative to the horizon at the beginning.
After each navigation step, we update the agent pose by the accumulation of discrete actions.

At the start, the agent is driven to perceive its panoramic surroundings. This is achieved by executing four sequential 90-degree rotations, enabling the agent to learn and represent the scene comprehensively. It facilitates scene exploration and effectively finds the target object.
After the initial action program, the agent sequentially executes planned task primitives.

\subsection{Perception}

The perception system starts with a $300\times300$ egocentric RGB frame. The field of view of the deployed camera is 60 degrees. We leverage three neural nets to estimate pixel-wise instance segmentation, depth, and affordances in the same resolution. Pixels are projected into the 3D space as point clouds, then localized in the allocentric map. We determine grids with more than 500 points are visible in the view. The representations of visible grids are learned by the aggregated semantic label of point clouds. Details of the perception and scene modeling are in Sec.~3.1 and Sec.~3.2, the main text.

\begin{figure}[t]
    \centering
    \includegraphics[width = 0.95\linewidth]{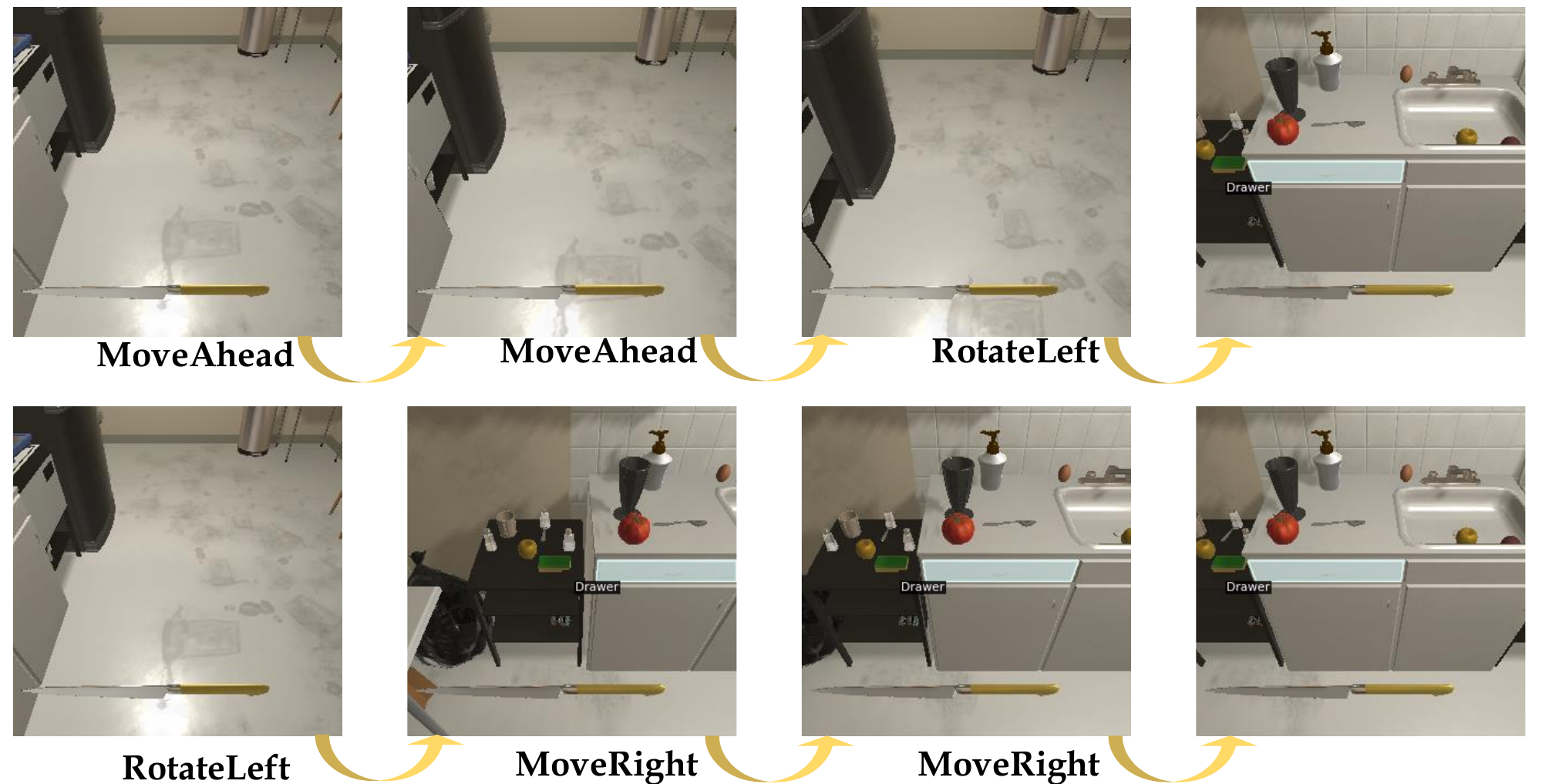}
    \caption{Different trajectories in navigation. The top trajectory is from the original ALFRED data. The target object(drawer in white) is not visible because of a wrong view direction in moving steps, leading to learning ambiguity. The bottom trajectory is generated by us. We adjust agent rotation at the start of the fine control to force object visibility.}
    \label{fig:supp_data}
\end{figure}

\subsection{Action Control}

We employ dual-level coarse-to-fine action controls. The coarse control is based on the global scene map but the fine control is based on the local egocentric view. 

Breadth-First Search (BFS) is used to determine the coarse actions. It plans the shortest navigable trajectory from the current position to the destination. Notably, the destination is a random point if the object is never found. Once it is found, the BFS destination is the localized semantics and its neighbor grids within 1 meter. 

Neural policy is used to determine the fine actions. They are trained by imitating planned actions from an expert. The neural action space includes all navigation actions and an \texttt{Interact} action. 
At the start of the neural control, we adjust agent rotation referring to the object direction in the localized scene map. This ensures object visibility for the policy network.
The neural policy control steps navigation actions and terminates until predicting the \texttt{Interact} action. Then the specific interactive action determined by the verb primitive is executed.
The fine neural control holds two insights, \ie object-oriented and short-horizon. First, we adjust the agent pose conditioned on the target object state, which is always visible in the fine adjustment. This reduces the ambiguity in object manipulation. Second, the dual-level controls naturally decrease the length of learning steps. Our neural policy is applied in short-horizon processes and costs fewer training steps.

After the moving steps, the agent performs specific interactions defined in~\cref{sec:intro_alfred}. A notable case is to interact with receptacles.
Our affordance module detects whether the receptacle is openable to plan the necessity of an open step.

More details of the control are in Sec.~3.3 of the main text.

\section{Training Trajectories}
We illustrate different training trajectories in~\cref{fig:supp_data}, including the trajectory from ALFRED and those generated by us. 
We can find an intriguing property of different trajectories.
The agent following the ALFRED trajectory may have a wrong view in the marching stage and the object is not visible in the camera view. This phenomenon causes ambiguity in actions and makes it hard to learn. In contrast, our method is object-oriented. We adjust agent rotation using the localized yaw angle at the start and force the target object to always be visible in steps for learning. 
The object-oriented trajectory is more learnable and predictable than default ALFRED data.

\section{Additional Experiments}

\subsection{Analysis on Different Task Types}
For a more comprehensive evaluation, we report DISCO's performances on different tasks in \cref{tab:result_per_type}.
The stack task (\eg put a spoon in a mug then put the mug in a cabinet) turns out to be the most challenging for DISCO, as seen the 27.5\% success rate in unseen scenes. 
It involves more objects to be manipulated and requires many spatial relationships to be achieved in the final state.
DISCO also has subpar performances on heat and cool tasks, namely 54.4\% and 52.9\% success rates in unseen scenes. These two types of tasks require agents to manipulate microwaves or fridges to change object states. They turn out to be much more challenging than the common pick-place. It reveals there are also some potential improvements for DISCO to be more skilled in complex manipulations.

\begin{table}[t]
    \centering
\begin{minipage}[t]{0.55\textwidth}
    \centering
    \caption{Results per task type.}
    \label{tab:result_per_type}
    \resizebox{\textwidth}{!}{
          \begin{tabular}{l|cc|cc}
    \toprule
         &\multicolumn{2}{c|}{\bf Valid Seen} &\multicolumn{2}{c}{\bf Valid Unseen}\\    
         & SR & GC & SR & GC \\
         \midrule
    Look light   & 71.2 & 76.6 & 79.7 & 84.9 \\
    Pick \& Place& 66.9 & 66.9 & 53.0 & 53.0 \\
    Place two    & 62.1 & 73.1 & 38.3 & 56.7 \\
    Stack        & 40.8 & 51.4 & 27.5 & 32.4 \\
    Heat         & 46.7 & 60.3 & 54.4 & 70.6 \\
    Cool         & 42.0 & 52.7 & 52.9 & 71.2 \\
    Clean        & 72.3 & 79.9 & 61.1 & 73.0 \\
    All          & 57.3 & 63.9 & 55.0 & 65.5 \\
    \bottomrule
  \end{tabular}

    } 
\end{minipage}
\end{table}
\begin{table}[t]
\begin{center}
\caption{Failure analysis.}
\label{tab:fail}
\resizebox{0.6 \linewidth}{!}{
  \begin{tabular}{l|c|c}
    \toprule
    &{\bf Valid Seen} & {\bf Valid Unseen}\\  
    \midrule  
    Success Rate            & 57.3  & 55.0 \\
    \midrule
    Language error          &  18.3 & 19.3 \\
    Object not found        &  5.0  & 3.0\\
    Navigation collision    &  8.2  & 14.0\\
    Interaction failure     &  10.3 & 7.4\\
    Others                  &  0.9  & 1.3\\
    \bottomrule
  \end{tabular}
}
\end{center}
\end{table}

\subsection{Failure Analysis}
We conduct a comprehensive analysis of DISCO's failures in this section. 
Quantitative results are reported in~\cref{tab:fail}.
DISCO can achieve 55\%+ success rate in both seen and unseen scenes, but there is still potential room for improvement.
Language misunderstanding errors contribute to the largest proportion of failures, nearly one-fifth. It suggests the necessity of stronger language models in embodied applications.
In some episodes, the agent fails to find the target object. A possible reason is that the object may be located in a closed receptacle and is not visible in the wild. Active exploration with commonsense knowledge is required to address this issue.
Other common failures include navigation collision and interaction failure, approximately 10\% for each.

\begin{table}[t]
\begin{center}
\caption{Ablation study of differentiable representations.}
\label{tab:abl_diff_optim}
\resizebox{0.65 \linewidth}{!}{
  \begin{tabular}{c|c|c}
    \toprule
    \hspace{10pt}Representation\hspace{10pt} & Num. Optim. Steps &  Success Rate\\
    \midrule  
    Differentiable  & 5  & 53.6 \\
    Differentiable  & 10 & 55.0 \\
    Differentiable  & 15 & 54.7 \\
    Cell   & - & 42.7 \\
    \bottomrule
  \end{tabular}
}
\end{center}
\end{table}

\subsection{Ablation Study of Differentiable Representations}

The quality of differentiable representations is affected by its optimization process. We provide an additional ablation study of the optimization steps of differentiable features. We use 10 optimization steps by default but also report results using 5 and 15 optimization steps in Tab.~\ref{tab:abl_diff_optim}. 
In \textit{valid unseen} set, DISCO achieves 53.6\%, 55.0\%, and 54.7\% success rates using 5, 10, and 15 optimization steps respectively. And they all greatly outperform the cell representation baseline of 42.7\% success rate. It validates both the effectiveness and robustness of differentiable representations.

\subsection{Application in Interaction Exploration}
To evaluate the generalized ability of DISCO, we conduct additional experiments in an interaction exploration task following~\cite{nagarajan2020learning}. The interaction exploration task is to quickly discover what objects can be interacted and how to interact in an embodied environment. It puts requirements on scene understanding, object interaction localization, and action planning. 

We apply DISCO in interaction exploration tasks. Same to the implementation in ALFRED tasks, our perception module and differentiable scene representations can be seamlessly integrated to understand environments in interaction exploration. In the action planning stage, we localize interactable objects using the affordance module and select the nearest object as the target. Our dual-level control drives the agent to perform all possible interactions with the target object.

Following~\cite{nagarajan2020learning}, we simulate in \textit{kitchen} scenes in AI2-THOR~\cite{ai2thor}. There are 20 scenes in the training set and 5 scenes in the testing set respectively. The action space aligns ALFRED setting. The maximum time step is 1,024. We use two metrics to evaluate our method: \textbf{(1) Coverage}: the ratio of executed interactions to the maximum possible interactions. \textbf{(2) Precision}: the ratio of executed interactions to the number of attempts.

Results in the interaction exploration task are reported in Tab.~\ref{tab:intexp}. DISCO outperforms the IntExp~\cite{nagarajan2020learning} in both metrics, namely +4.0\% in Coverage and +5.3\% in Precision. It validates the generalization of DISCO.

\begin{table}[t]
\begin{center}
\caption{Results in interaction exploration.}
\label{tab:intexp}
\resizebox{0.4 \linewidth}{!}{
  \begin{tabular}{l|c|c}
    \toprule
    & {\bf Coverage} & {\bf Precision}\\  
    \midrule  
    IntExp~\cite{nagarajan2020learning}  & 22.2 & 8.5\\
    DISCO(Ours)   & {\bf 26.2} & {\bf 13.8}\\
    \bottomrule
  \end{tabular}
}
\end{center}
\end{table}

\begin{figure*}[t]
    \centering
    \includegraphics[width = 0.95\linewidth]{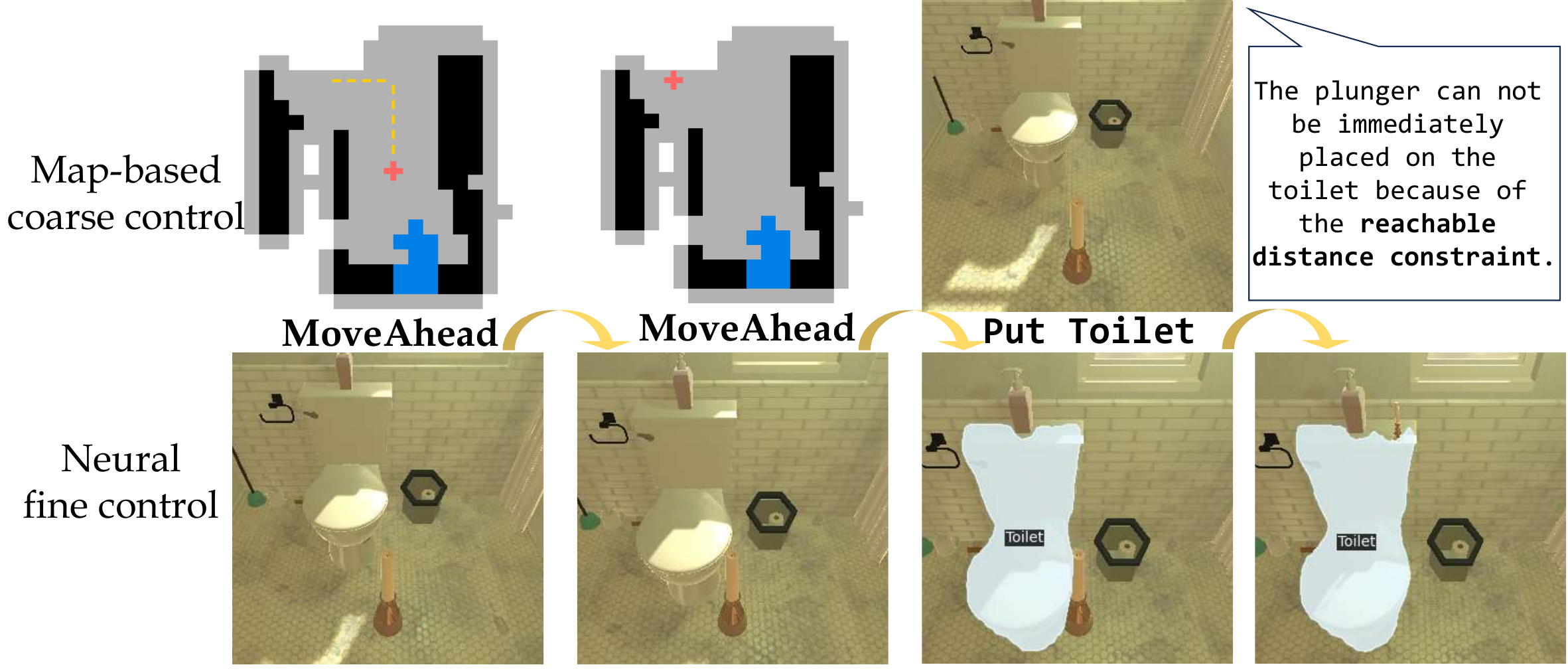}
    \includegraphics[width = 0.95\linewidth]{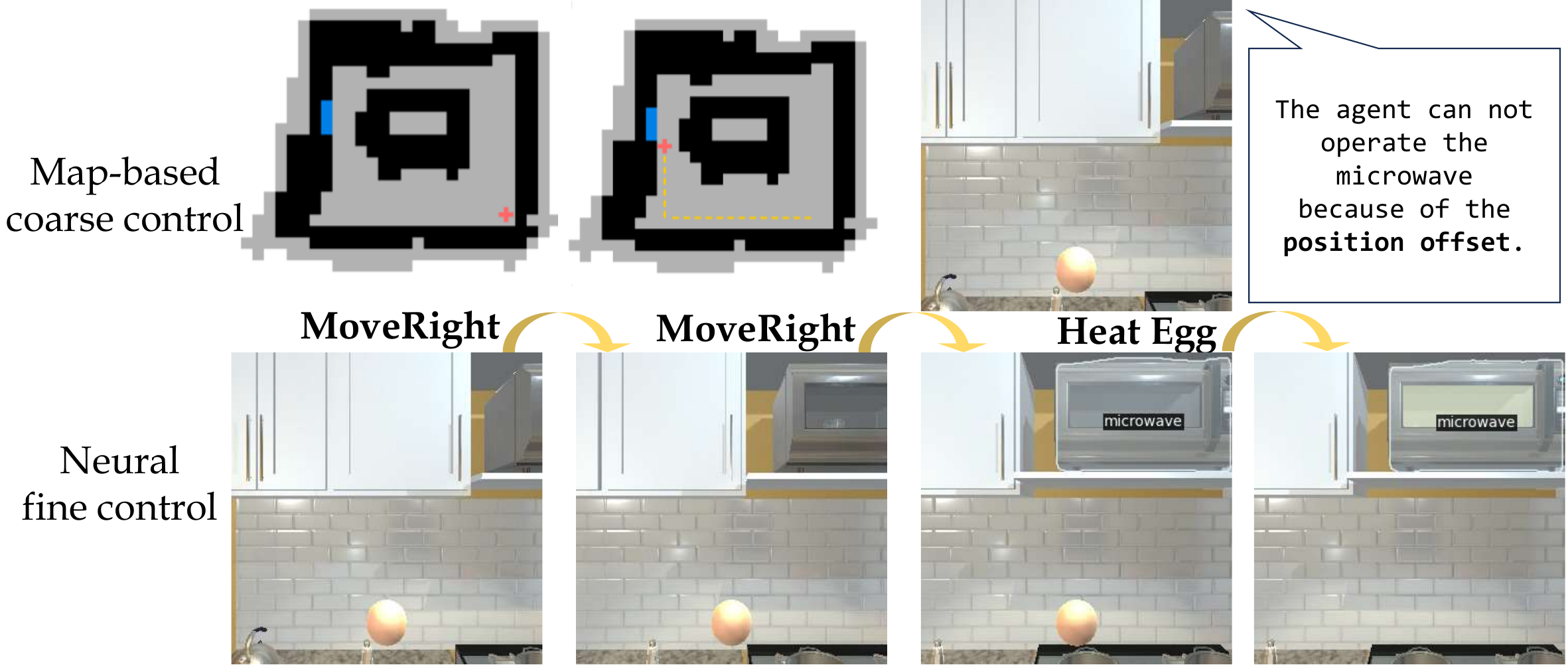}
    \caption{Qualitative running cases. After coarse navigation, the agent may suffer from reachable distance constraint (upper) or position offset (bottom) which causes interaction failure. Fine actions perform self-adjustment to address the issues.}
    \label{fig:supp_case}
\end{figure*}

\subsection{Qualitative Results}

We provide more qualitative results in~\cref{fig:supp_case} to demonstrate the effectiveness of our method, as a supplement to Sec. 4.6 of the main text. In these cases, the agent can not interact with objects after the coarse control process, because of reachable distance constraint or position offset. Our dual-level controls solve the issue.
 
\clearpage

\end{document}